%% file: root.tex
\documentclass[letterpaper, 10 pt, conference]{ieeeconf}  

\IEEEoverridecommandlockouts                              

\usepackage[font=footnotesize]{caption}
\usepackage{subcaption} 
\usepackage[table]{xcolor} 
\usepackage{etoolbox}
\usepackage{booktabs}
\usepackage{graphicx}
\usepackage{amsmath,amssymb,amsfonts}
\usepackage{hyperref}
\usepackage{cite}
\usepackage{arydshln}
\usepackage{tikz}
\usetikzlibrary{shapes,arrows}
\usetikzlibrary{arrows.meta}
\usetikzlibrary{angles,quotes}
\usetikzlibrary{3d}

\title{\LARGE \bf
  Intercepting an Agile Target with Net-Carrying Drones using Competitive Multi-Agent Reinforcement Learning
}

\author{\authorblockN{Timothée Gavin\authorrefmark{1}\authorrefmark{2}\authorrefmark{3}, and Murat Bronz\authorrefmark{2} } \authorblockA{\authorrefmark{1}\textit{IAS}, \textit{Thales LAS}, Rungis, France}
\authorblockA{\authorrefmark{2}\textit{Dynamic Systems, OPTIM}, \textit{Fédération ENAC ISAE-SUPAERO ONERA, Université de Toulouse}, Toulouse, France}
\authorblockA{\authorrefmark{3}\textit{RIS, } \textit{LAAS-CNRS} Toulouse, France}
\authorblockA{timothee.gavin@thalesgroup.fr , murat.bronz@enac.fr}}

\newcommand{\BibTeX}{\mathrm B\kern-.05em{\sc i\kern-.025em b}\kern-.08em\TeX}
\definecolor{pursuerblue}{HTML}{4c72b0}

\definecolor{evaderorange}{HTML}{dd8452}

\usepackage{colortbl}
\usepackage{pgf}

\newcommand{\gradientcelld}[8]{%
  \xdef\lowvalx{#2}%
  \xdef\midvalx{#3}%
  \xdef\maxvalx{#4}%
  \xdef\lowcolx{#5}%
  \xdef\midcolx{#6}%
  \xdef\highcolx{#7}%
  \xdef\opacityx{#8}%
  \ifdimcomp{#1pt}{>}{\maxvalx pt}{%
    \begingroup\setlength{\fboxsep}{0pt}\cellcolor{\highcolx!100.0!\midcolx!\opacityx}#1\endgroup%
  }{%
    \ifdimcomp{#1pt}{<}{\midvalx pt}{%
      \ifdimcomp{#1pt}{<}{\lowvalx pt}{%
        \begingroup\setlength{\fboxsep}{0pt}\cellcolor{\midcolx!0.0!\lowcolx!\opacityx}#1\endgroup%
      }{%
        \pgfmathparse{int(round(100*(#1/(\midvalx-\lowvalx))-(\lowvalx*(100/(\midvalx-\lowvalx)))))}%
        \xdef\tempa{\pgfmathresult}%
        \begingroup\setlength{\fboxsep}{0pt}\cellcolor{\midcolx!\tempa!\lowcolx!\opacityx}#1\endgroup%
      }%
    }{%
      \pgfmathparse{int(round(100*(#1/(\maxvalx-\midvalx))-(\midvalx*(100/(\maxvalx-\midvalx)))))}%
      \xdef\tempb{\pgfmathresult}%
      \begingroup\setlength{\fboxsep}{0pt}\cellcolor{\highcolx!\tempb!\midcolx!\opacityx}#1\endgroup%
    }%
  }%
}

\definecolor{heatbad}{HTML}{D65F5F}
\definecolor{heatmid}{HTML}{F2E394}
\definecolor{heatgood}{HTML}{5AA469}
\xdef\opacity{65}

\newcommand{\gH}[1]{\gradientcelld{#1}{0.0}{50.0}{100.0}{heatbad}{heatmid}{heatgood}{\opacity}}
\newcommand{\gL}[1]{\gradientcelld{#1}{0.0}{30.0}{100.0}{heatgood}{heatmid}{heatbad}{\opacity}} 
\newcommand{\gT}[1]{\gradientcelld{#1}{2.82}{6.41}{10.00}{heatgood}{heatmid}{heatbad}{\opacity}}

\newcommand{\gLs}[1]{{\scriptsize \gL{#1}}}
\newcommand{\gTs}[2]{%
  \xdef\lowvalx{2.82}%
  \xdef\midvalx{6.41}%
  \xdef\maxvalx{10.00}%
  \xdef\lowcolx{heatgood}%
  \xdef\midcolx{heatmid}%
  \xdef\highcolx{heatbad}%
  \xdef\opacityx{\opacity}%
  \ifdimcomp{#1pt}{>}{\maxvalx pt}{%
    \begingroup\setlength{\fboxsep}{0pt}\cellcolor{\highcolx!100.0!\midcolx!\opacityx}{\scriptsize #2}\endgroup%
  }{%
    \ifdimcomp{#1pt}{<}{\midvalx pt}{%
      \ifdimcomp{#1pt}{<}{\lowvalx pt}{%
        \begingroup\setlength{\fboxsep}{0pt}\cellcolor{\midcolx!0.0!\lowcolx!\opacityx}{\scriptsize #2}\endgroup%
      }{%
        \pgfmathparse{int(round(100*(#1/(\midvalx-\lowvalx))-(\lowvalx*(100/(\midvalx-\lowvalx)))))}%
        \xdef\tempa{\pgfmathresult}%
        \begingroup\setlength{\fboxsep}{0pt}\cellcolor{\midcolx!\tempa!\lowcolx!\opacityx}{\scriptsize #2}\endgroup%
      }%
    }{%
      \pgfmathparse{int(round(100*(#1/(\maxvalx-\midvalx))-(\midvalx*(100/(\maxvalx-\midvalx)))))}%
      \xdef\tempb{\pgfmathresult}%
      \begingroup\setlength{\fboxsep}{0pt}\cellcolor{\highcolx!\tempb!\midcolx!\opacityx}{\scriptsize #2}\endgroup%
    }%
  }%
}

\newcommand{\GradientLegendStrip}[9][4.2cm]{%
  \begingroup
  \pgfmathsetmacro{\midpos}{(#3-#2)/(#4-#2)}%
  \begin{tikzpicture}[x=#1,y=1ex,baseline=(legendlabel.base)]
    \node[anchor=east] (legendlabel) at (0,0) {\scriptsize #9};

    \foreach \i in {0,...,99}{%
      \pgfmathsetmacro{\v}{#2 + (\i+0.5)*(#4-#2)/100}%
      \ifdimcomp{\v pt}{<}{#3 pt}{%
        \pgfmathsetmacro{\p}{100*(\v-#2)/(#3-#2)}%
        \edef\legendcolor{#6!\p!#5!#8}%
      }{%
        \pgfmathsetmacro{\p}{100*(\v-#3)/(#4-#3)}%
        \edef\legendcolor{#7!\p!#6!#8}%
      }%
      \fill[\legendcolor] ({\i/100},0) rectangle ({(\i+1)/100},1);
    }%

    \draw[black!35] (0,0) rectangle (1,1);

    \node[anchor=north] at (0,0.30) {\tiny #2};
    \node[anchor=north] at (\midpos,0.30) {\tiny #3};
    \node[anchor=north] at (1,0.30) {\tiny #4};
  \end{tikzpicture}%
  \endgroup
}
\newcolumntype{K}{!{\color{white}\ }r}

\begin{document}

\maketitle
\thispagestyle{empty}
\pagestyle{empty}

\begin{abstract}

  This article presents a solution to intercept an agile drone by a team of agile drone carrying catching nets.
  We formulate the problem as a competitive Multi-Agent Reinforcement Learning (MARL) task. 
  To address the problem of non-stationarity and catastrophic forgetting of agents overfitting to the current opponent strategy, we train the pursuers and the evader using Multi-Agent Proximal Policy Optimization (MAPPO) with Prioritized Fictitious Self Play (PFSP).
  We train the agents in a high-fidelity simulator using low-level control commands, collective thrust and body rates (CTBR), to achieve agile flights for both the pursuers and the evader.
  We compare the performance of the trained policies in terms of catch rate, time to catch and crash rates, against heuristic baselines and show that our solution outperforms them.
  Ablation studies show that PFSP lead to more robust policies that can adapt to different opponent strategies, and that a low-level control commands are crucial for learning performing strategies in the pursuit-evasion task.
  Finally, a qualitative analysis of the learned behaviours highlights the emergence of cooperative tactics among the pursuers.

\end{abstract}



\section{Introduction}

Intercepting agile aerial targets with autonomous drones is a key challenge in robotics and security. 
As Unmanned Aerial Vehicle (UAV) incursions into restricted airspace become more common, applications such as airspace protection, infrastructure security, and event safety demand solutions that can neutralize unauthorized drones with minimal collateral damage \cite{Park2021}.
Fleets of interceptor drones with capture-nets are a promising option, but their success depends on advanced control and coordination against evasive targets.

Classical interception methods rely on accurate models, preplanned strategies, or predictable target behaviour \cite{Yanushevsky2018}. 
However, modern quad-rotor drones can perform highly dynamic manoeuvres, and will actively evade capture, making classical methods ineffective \cite{Chung2011}.

Recent advances in deep reinforcement learning (RL) have shown that drones can learn agile flight behaviours directly from interaction with the environment.
RL-trained policies achieved superhuman performance in drone racing, where agents navigate challenging courses with dynamic manoeuvres \cite{Kaufmann2023champion}. 
However, drone racing typically involves static or predictable gates, whereas interception requires agents to respond to adversarial targets that actively evade capture.
At the same time, Multi-Agent RL (MARL) has demonstrated remarkable success in competitive environments, particularly in complex games \cite{Silver2017, OpenAI2019}.
\begin{figure}[t]
  \centering
  \includegraphics[width=0.9\linewidth]{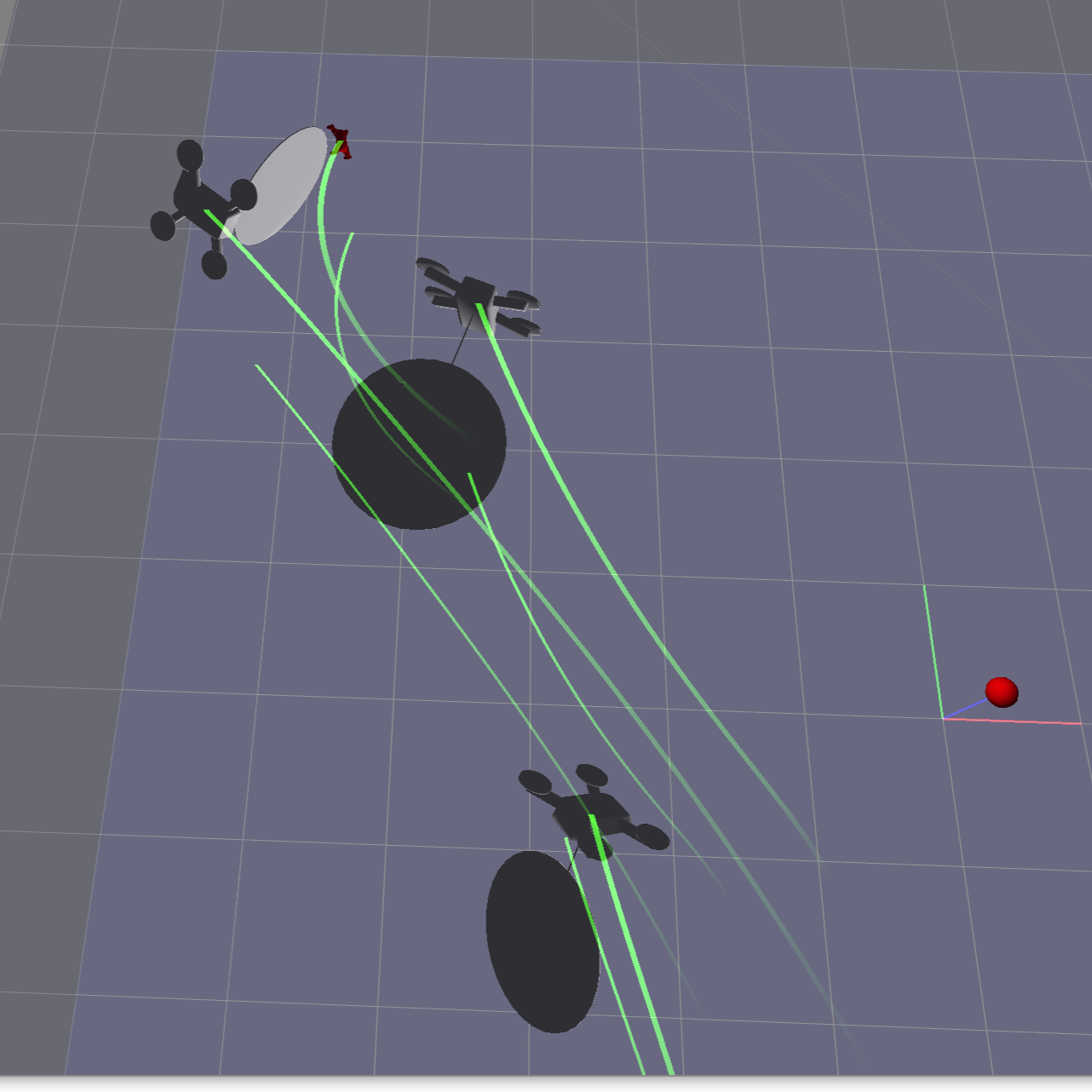} 
  \caption{A multi-agent competitive reinforcement learning approach to train both a team of pursuers and an evader drone for agile pursuit-evasion using catching-nets. Both pursuers and evader learn low-level control policies that enable them to perform dynamic manoeuvres in a high-fidelity simulator.}
  \label{fig:catch_it}
\end{figure}

In this work, we formulate agile drone interception between a team of net-carrying pursuers and an evasive target as a competitive MARL task.
The main contributions of this paper with respect to the literature are:
\begin{itemize}
  \item A competitive MARL framework for cooperative interception of an agile evader by multiple pursuer drones using low-level control.
  \item A Prioritized Fictitious Self-Play training scheme for both the pursuers and the evader to learn robust policies against varied opponents.
  \item Extensive simulation results demonstrating stronger performance than standard baselines.
\end{itemize}

The remainder of the paper is organized as follows. 
Section \ref{sec:related_work} reviews related work on interception using teams of drones.
Section \ref{sec:background} provides background on reinforcement learning and competitive self-play.
Section \ref{sec:methodology} details our training methodology.
Section \ref{sec:experiments} presents experimental results and comparisons.
Section \ref{sec:conclusions} concludes and discusses future directions.

\section{Related Work}\label{sec:related_work}


Agile flights in multi-rotors drones is typically characterized by the ability to perform large-angle manoeuvres, sustain high linear and angular accelerations, maintain precise control near dynamic limits and do so reliably in real-time, often in complex and cluttered environments.
Traditional guidance laws for interception, such as Proportional Navigation (PN), are computationally simple and widely used in missile guidance.
But, they typically assume predictable target manoeuvres and are less effective against highly agile or adversarial evaders. 
PN was recently adapted for catching manoeuvring quadrotors \cite{pliska2024towards}, yet still rely on predictable target models.
This issue can be mitigated by limiting the evader's escape options using multiple pursuers, 
but this requires additional mechanisms such as target-pursuer allocation \cite{asadi2021optimal}, inter-pursuers collision avoidance \cite{zhu2017distributed} or a geometric coordination layer \cite{zhou2016cooperative} to ensure proper deconfliction and coverage of the evader's escape routes.

Learning-based methods, particularly MARL, offer an alternative for adversarial multi-UAV interception.
In \cite{Xiao2024}, the authors introduces a co-evolution framework that uses RL to train both the pursuers and the evader in the pursuit-evasion task.
However, like most RL approaches \cite{Zhang2023, zhang2023dacoop}, they rely on high-level control inputs, in this case velocity commands, and simplified quadrotor dynamics, which limits the agility of the learned behaviours.
In contrast, \cite{Chen2024} uses RL for low-level control of agile pursuers, however they assume fixed evader behaviours and only consider the pursuit aspect.
Recently, \cite{roncero2025learned} and \cite{gavinagile} both combined a competitive RL framework and agile control inputs, but they limit their study to the 1v1 pursuit-evasion case.
Overall, while RL has shown promise in the multi-pursuer interception task, current research either overlooks the adversarial element of a learning evader or typically lacks highly dynamic flight capabilities in both the pursuers and the evader.



\section{Background}\label{sec:background}
\subsection{Multi-Agent Reinforcement Learning}\label{sec:background_marl}

Reinforcement Learning is a type of machine learning for sequential decision-making.
In a \textit{rollout} phase, an \textit{agent} interacts with an uncertain \textit{environment} which provides it with \textit{partial observations} of its \textit{state}, takes a series of \textit{actions} following a \textit{policy} and receives a scalar feedback in the form of \textit{rewards}.
These sequences of observe-act-reward, repeated over time, form the \textit{rollouts}.
The collected rollouts are then used to update the policy in a learning phase, which will then be employed in the rollout phase of the next training iteration.
The goal of the agent is to learn a policy that maximizes the expected cumulative reward over time.

Multi-Agent Reinforcement Learning (MARL) extends RL to scenarios with multiple agents interacting in a shared environment. 
MARL suffers from the curse of dimensionality and non-stationarity, as the environment dynamics change as other agents learn and adapt their policies.
Recent works in MARL adopted centralized training with decentralized execution (CTDE) \cite{Lowe2017}, where agents have access to global information during training but operate based on local observations during execution. In competitive settings, this alleviates non-stationarity by allowing the agents to access the state and actions of their opponents during training.

\subsection{Self-Play}\label{sec:background_self_play}


Self-Play (SP) is a training method where an agent plays against oneself without any direct supervision. 
It yields excellent performance in challenging tasks such as Go, Poker \cite{Silver2017}, and competitive video games like Dota 2 and StarCraft II \cite{OpenAI2019, vinyals2019grandmaster}.
In practice, each agent periodically freezes its policy while the other agent adapt to that frozen policy. Then the roles swap. 
But, it famously suffers from the problem of overfitting to a single opponent and forgetting how to win against past versions.
Fictitious Self-Play is one solution to solve this problem. Each agent learns the best response to the empirical frequency distribution of the opponent’s past strategies.
However, using FSP, the agents waste time playing against past versions of themselves that are too weak or too strong to provide a meaningful learning signal.
Prioritized Fictitious Self-Play (PFSP) \cite{vinyals2019grandmaster, OpenAI2019} addresses this issue by improving the diversity of the opponents' policies the agents face. 
By sampling opponent policies from a pool according to a performance-based distribution, it ensures continual exposure to past strategies while prioritizing more challenging opponents.

\section{Methodology}\label{sec:methodology}

\begin{figure*}[!htbp]
  \centering
  \includegraphics[width=\textwidth, clip, trim=0 0 0 0]{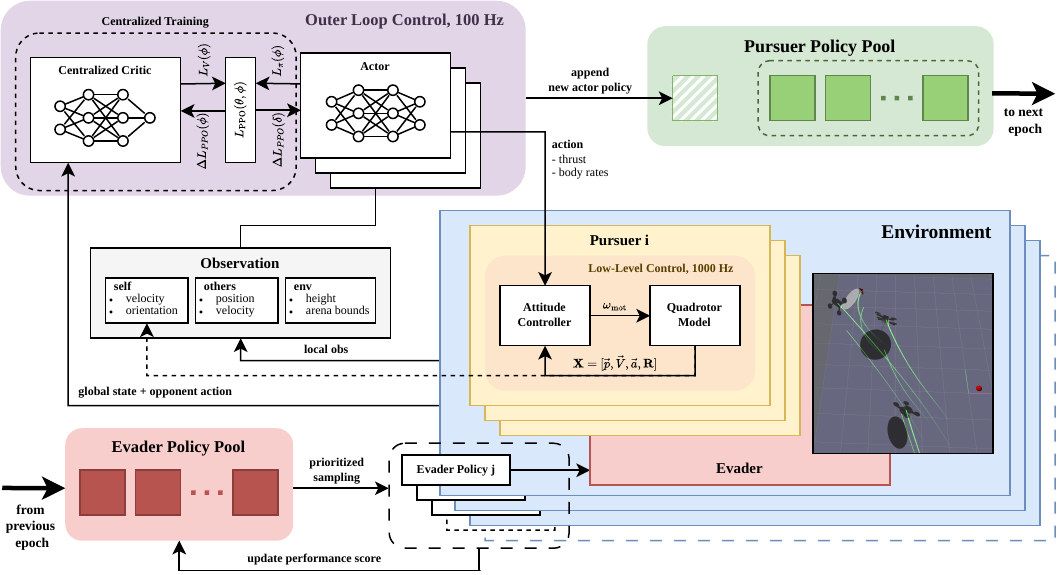}
  \caption{
    Overview of the training method from the pursuers' perspective: 
    In each epoch, one actor network per team is trained with MAPPO and its parameters are shared across all same-team agents for decentralized execution. 
    The environment is vectorized to allow batched rollouts.
    Each environment episode is played against a different frozen opponent policy sampled from the opponent pool using a performance-weighted distribution. 
    Training uses CTDE: the centralized critic observes the global state and opponent actions, while the actor uses only local observations and outputs thrust/body-rate commands tracked by the low-level controller. 
    When the current policy exceeds a performance threshold against the full opponent pool, a snapshot is frozen and stored in the team's growing policy pool, and training switches to the other team.
  }
  \label{fig:method}
\end{figure*}

\subsection{Problem Formulation}\label{sec:methodology_problem_formulation}

We study a pursuit–evasion scenario with quadrotors, a team of $N$ \emph{pursuers} and a single \emph{evader} with identical manoeuvring capabilities, operating in an obstacle-free rectangular arena.
The pursuers seek to catch the evader quickly, while the evader tries to avoid capture. 
At the beginning of each episode, the agents’ initial positions are drawn uniformly at random inside the arena.  
Capture occurs when the evader’s centre comes within a \emph{capture distance} of any pursuer’s rigid, circular net of radius \(R\), which is mounted below each pursuer and aligned with its body frame.
In such settings, the evader quickly learns to exploit the pursuers’ tendency to avoid boundary violations by flying close to the arena walls.
To discourage this behaviour and promote agile evasive flight in the central region, a buffer zone is added along each wall. 
Only the evader is penalized for entering this zone, which restricts their movement to a smaller volume in the centre of the arena.

Target detection, state estimation, and trajectory prediction are not addressed in this work. We acknowledge that, together with the obstacle-free
arena, and the constrained area for the evader, these are strong assumptions that affect the behaviour of the agents and the practical relevance of the results, but we reserve the relaxation of these assumptions for future work.  

\subsection{Observation and Action Space}\label{sec:methodology_observation_and_action_space}
We denote $\mathcal{N_{\mathrm P}} = {1, \ldots, N}$ the set of $N>1$ pursuers and $e$ the single evader.
We denote the position, velocity, and rotation matrix of agent $i\in \mathcal{N_{\mathrm P}} \cup \{e\} $, in world coordinates as $\mathbf{p}_i$, $\mathbf{v}_i$, and $\mathbf{R}_i$ respectively.
For an agent $j\in \mathcal{N_{\mathrm P}} \cup \{e\} $, we denote its relative position and velocity in agent $i$'s body frame as $\mathbf{p}^{i}_j = \mathbf{R}_i^\top(\mathbf{p}_j - \mathbf{p}_i)$ and $\mathbf{v}^{i}_j = \mathbf{R}_i^\top(\mathbf{v}_j - \mathbf{v}_i)$.

For each agent $i \in \mathcal{N_{\mathrm P}} \cup \{e\} $, its observation $\mathbf{o}_i$ is composed of the self-state observation $\mathbf{o}^{\mathrm self}_i$, the observation of the other agents $\mathbf{o}^{\mathrm others}_i$ (including the evader in the case of pursuers), and the observation of the arena bounds and the ground $\mathbf{o}^{\mathrm env}_i$.
The self-state observation is $\mathbf{o}^{\mathrm self}_i=[\mathbf{R}_i^\top\mathbf{v}_i, \text{vec}(\mathbf{R}_i)]$ containing the agent's linear velocity in body frame $\mathbf{R}_i^\top\mathbf{v}_i$, and its rotation matrix $\mathbf{R}_i$, with $\text{vec}(\cdot)$ being the flattening function.
The observation of the other agents is $\mathbf{o}^{\mathrm others}_i=[\mathbf{p}^i_j, \mathbf{v}^i_j]_{j\in {\mathcal{N_{\mathrm P}}\cup\{e\}\backslash\{i\}}}$ containing the position and velocity of each agent $j$ relative to the agent expressed in the agent body frame.
Finally, the observation of the arena bounds and the ground is $\mathbf{o}^{\mathrm env}_i=\{z_i\}\cup\{\mathbf{d}_m\}_{m=1}^M$ and is composed of the altitude of the agent $z_i$ and a set of distance to arena bounds along $M$ rays uniformly distributed in the horizontal plane.
We normalize the observations before passing them to the neural network.

The control policies are trained using Multi-Agent Proximal Policy Optimization (PPO) \cite{Yu2022}. This Actor-Critic method uses two neural networks for each team of agents: a policy network and a value network and follows the CTDE paradigm.
Each team has a single shared policy network that is executed independently by each agent using only its local observation, generating, for each agent, an action $\mathbf{a}_i$, composed of a collective thrust $\mathbf{a}^{th}_i$ and body rates $\mathbf{a}^\omega_i$.
And a single centralized value network used only during training that receives privileged global information about the opponent team's state, which is not available to the policy network. 
This alleviates the non-stationarity of the environment due to the simultaneous learning of all agents \cite{Lowe2017}.
The input of the value network of each team is the concatenation of the exact position, velocity, and rotation matrix of each agent, as well as the actions taken by the opponent(s) at this time step. 
This input is normalized before being fed to the neural network.

\subsection{Rewards}\label{sec:methodology_rewards}
The reward of the pursuers $r^{\text{P}}_p$ ($p \in \mathcal{N_{\mathrm P}}$) and the evader $r^{\text{E}}$ are given by:
\begin{align*}
  r^{\mathrm{P}}_p & = \phantom{-}r^{\mathrm{catch}} - r^{\mathrm{step}} - r^{\mathrm{coll}}_{p,\mathrm{P}} - r^{\mathrm{coll}}_{p,\mathrm{E}} - r^{\mathrm{fail}}_p - r^{\mathrm{cmd}}_p - r^{\mathrm{dist}}_p, \\
  r^{\mathrm{E}}  & =             -r^{\mathrm{catch}} + r^{\mathrm{step}} - r^{\mathrm{coll}}_{e,\mathrm{P}} - r^{\mathrm{fail}}_e - r^{\mathrm{cmd}}_e - r^{\mathrm{bnd}}.
\end{align*}
in which $r^{\mathrm{catch}}$ rewards all the pursuers when one catches the evader,
$r^{\mathrm{fail}}_i$ penalizes any agent for crashing or going out of bounds, 
$r^{\mathrm{coll}}_{p,\mathrm{P}}$ penalizes any pursuer for colliding with the body of another pursuer, 
$r^{\mathrm{coll}}_{p,\mathrm{E}}$ penalizes collisions between any pursuer body and the evader body,
$r^{\mathrm{step}}$ rewards the evader for surviving each time step,
and $r^{\mathrm{cmd}}$ discourages dynamically infeasible commands.
Instead of terminating the episode upon collision between a pursuer body and the evader, we apply a soft continuous penalty $r^{\mathrm{coll}}_{p,\mathrm{E}} = r^{\mathrm{coll}}_{e,\mathrm{P}}$ to both agents, allowing for gradual learning of collision avoidance while maintaining focus on the primary task of catching with the net.
However, for inter-pursuer collisions, our experiments showed that soft penalties slow down the learning of effective spreading tactics.
So we terminate the episode for inter-pursuer collisions and apply an important penalty $r^{\mathrm{coll}}_{p,\mathrm{P}}$.
We still terminate the episode if any agent goes out of bounds or crashes on the ground and apply a hard penalty $r^{\mathrm{crash}}_i$.
Yet, neither the evader nor the pursuers receive a reward when the opponent reaches a failure state, encouraging actual pursuit–evasion behaviour rather than strategies that force the opponent to crash.
Additionally, the reward shaping terms
$r^{\mathrm{dist}}_p$ penalizes the pursuer for being far from the evader to encourage pursuit,
and $r^{\mathrm{bnd}}$ penalizes the evader when approaching the arena bounds to discourage exploiting the boundaries.

Specifically, the reward terms are:
\begin{align*}
  r^{\rm catch} &= \lambda_{\rm catch} \cdot \mathbf 1_{\rm catch}, 
  &
r^{\mathrm step} &= \lambda_{\mathrm step}, 
  \\
  r^{\mathrm{coll}}_{p,\mathrm{E}} &= \lambda_{{\mathrm{coll}}{\mathrm{PE}}} \cdot \mathbf 1_{\mathrm{contact}}^{p,e},
  &
  r^{\mathrm{coll}}_{p,\mathrm{P}} &= \lambda_{{\mathrm{coll}}{\mathrm{PP}}} \cdot \mathbf 1_{\mathrm{contact}}^{p,\mathrm{P}},
  \\
  r^{\mathrm{fail}}_i &= \lambda_{\mathrm{fail}} \cdot \mathbf{1}_{\mathrm{fail}}^{i}, 
  &
  r^{\mathrm{cmd}}_i &= \lambda_{\mathrm{cmd}} \cdot \lVert\mathbf a_i^\omega\rVert.
\\
  r^{\rm dist}_p &= \lambda_{\rm dist}\cdot \bigl\lVert {\mathbf p}_e - {\mathbf c}^{\rm net}_p\bigr\rVert_2, 
  &
r^{\mathrm{bnd}} &= \phi_{\mathrm{bnd}}(d^{\mathrm{bnd}}),
\end{align*}
in which the indicator functions return $1$ when their condition is met :
$\mathbf 1_{\rm catch}$ when catching the evader,
$\mathbf 1_{\mathrm{contact}}^{p,e}$ in case of a contact between a pursuer $p$ and the evader,
$\mathbf 1_{\mathrm{contact}}^{p,\mathrm{P}}$ in case of a contact between a pursuer $p$ and any other pursuer,
and $\mathbf{1}_{\mathrm{fail}}^{i}$ when an agent $i$ reaches a failure state because of a ground crash or leaving the arena bounds.
$c^{\mathrm net}_p$ is the centre of the catching net of pursuer $p$,
and $\mathbf{a}_i^\omega$ are the commanded body rates of agent $i$.
$\phi_{\mathrm bnd}$ is a function that penalizes the evader for approaching the arena bounds, 
triggering under a set threshold and growing exponentially the shorter the distance to the arena bounds $d^{\mathrm bnd}$.
$\lambda_{\mathrm catch}$, $\lambda_{\mathrm dist}$, $\lambda_{\mathrm step}$, $\lambda_{{\mathrm col}\mathrm{PE}}$, $\lambda_{{\mathrm col}\mathrm{PP}}$, $\lambda_{\mathrm fail}$, $\lambda_{\mathrm cmd}$ are positive hyperparameters that balance the different reward terms and have been tuned to obtain the desired behaviour and listed in Table~\ref{tab:reward_coefficients}.

\begin{table}[b]
  \centering
  \caption{Reward coefficients.}
  \label{tab:reward_coefficients}
  \setlength{\tabcolsep}{6pt}
  \renewcommand{\arraystretch}{1.1}
  \begin{tabular}{@{}lclc@{}}
    \toprule
    \textbf{Coefficient} & \textbf{Value} & \textbf{Coefficient} & \textbf{Value} \\ 
    \midrule
    $\lambda_{\mathrm{catch}}$ & 10.0 & $\lambda_{\mathrm{step}}$ & 0.04 \\
    $\lambda_{{\mathrm{coll}}{\mathrm{PE}}}$ & 0.1 & $\lambda_{{\mathrm{coll}}{\mathrm{PP}}}$ & 10.0 \\
    $\lambda_{\mathrm{dist}}$ & 0.001 & $\lambda_{\mathrm{fail}}$ & 30.0 \\
    $\lambda_{\mathrm{cmd}}$ & 2e-04 &  &  \\
    \bottomrule
  \end{tabular}
\end{table}

\subsection{Training Method}\label{sec:methodology_training_method}


To avoid overfitting to the current opponent and becoming less effective against previously seen strategies, we use Prioritized Fictitious Self-Play \cite{vinyals2019grandmaster, OpenAI2019}.
The pursuers and the evader train in \emph{separate} loops and maintain a growing pool of past frozen policies.
During each training episode, each team's latest policy trains against a set of frozen opponent policies from the opponent’s pool, sampled according to a performance–weighted distribution.
When the performance score of an agent against the full current pool of opponent strategies is past a certain threshold, a snapshot of the team's current policy is frozen and stored in the growing policy pool.
This maintains exposure to past strategies while prioritizing the more challenging opponents.
Incidently, the opponent strategies being frozen during the training episode reduce the problem in each episode to a cooperative setting, and common MARL algorithms can apply.
An overview of the training method is illustrated in Figure~\ref{fig:method}.
Our agents play 50\% of the episodes against the latest opponent policy to ensure a continuous improvement by facing the most advanced opponent, and 50\% against the PFSP distribution.

The performance metric in our problem is not as simple as a binary catch/evade outcome. 
Drones can crash, which adds another possible outcome.
And an evader that can avoid interception for an extended time is a stronger opponent than one that is quickly caught.
During learning, pursuers should prioritize facing evader strategies that either make them often crash, or avoid capture for extended time and eventually reach timeout without being intercepted. However, weak evaders that crash a lot are less interesting and should be less sampled against.
On the contrary, the evader should prioritize facing pursuers strategies that either make it often crash, or that catch it quickly. But avoid those that crash frequently without providing a challenge.

According to this definition, we derive the following performance metrics for a pursuers' policy $i$ and an evader's policy $e$:
\begin{align}
  \rho_P(i, e) &= \begin{cases}
    1 - \frac{t}{\tau}, & \text{if $i$ catches $e$} \\
    0, & \text{if $i$ crashes, or if $e$ evades} \\
    1, & \text{if $e$ crashes} \\
  \end{cases}\label{eq:performance_metric_P} \\
  \rho_E(e, i) &= \left(1 - \rho_P(i, e)\right)\label{eq:performance_metric_E}
\end{align}
in which $t$ is the time step at which ends the episode, and $\tau$ is the maximum episode length. For a current training pursuer $i$, we sample the evader policy $e$ from the pool of past policies $\mathcal{E}$ with probability:
\begin{equation*}
  \frac{f\left(\rho_P(i, e)\right)}{\sum_{e' \in \mathcal{E}} f\left(\rho_P(i, e')\right)}
\end{equation*}
And similarly for a learning evader $e$, replacing $\rho_P(i, e)$ with $\rho_E(e, i)$. $f(x)=(1-x)^2$ is a weighting function that focuses PFSP on the hardest opponents. 

\subsection{Training Setup}\label{sec:methodology_training_setup}

Similarly to prior work \cite{gavinagile}, we use a high-fidelity simulator of the quadrotor dynamics, which include a low-level control architecture with collective thrust and body rates as inputs, and collisions between quadrotors, the elements of the arena and the net carried by the pursuers.
This simulator also include a high-level INDI controller \cite{smeur2016adaptive}, which allow us to give alternate high-level commands to the quadrotors, such as position, velocity, or acceleration commands. This will be used to compare our method with classical baselines in Section \ref{sec:experiments}.


Each policy network is a two‐layer multilayer perceptron with $256$ ReLU units per hidden layer.  
The output layer produces the mean and standard‐deviation of a multivariate Gaussian, followed by a \texttt{tanh} squashing to obtain bounded continuous actions.  
The value networks mirror this architecture but ends with a linear output.

Rollouts are generated in parallel across $1024$ environments.  
The entire pipeline, including the simulator, is written in Python using JAX \cite{Bradbury2018}, enabling just‐in‐time compilation and parallelized execution on accelerated hardware.
Running on a single machine equipped with an NVIDIA RTX~4090 (24\,GB VRAM), an AMD Ryzen 9 7950X3D (16~cores, 4.2\,GHz) and 128\,GB RAM, the system collects and processes approximately \(3.5\times10^{5}\) environment steps per second.
We train for a total of $5$ billion environment steps for each side ($10$ billion steps in total), which corresponds to roughly 5h30 of wall-clock training time.


\section{Experiments}\label{sec:experiments}

\subsection{Experiment Setup}\label{sec:experiments_experiments_setup}
Similar to prior work \cite{gavinagile}, the main comparison metrics are the catch rate of the pursuers, the evade rate of the evader, the time to catch and the crash rates. 
The catch rate is the percentage of episodes where pursuers catch the evader within 10 seconds without crashing.
The evade rate is the percentage where the evader avoids capture for 10 seconds.
Crash rates distinguish between pursuers crash rate and evader crash rate where either a pursuer or the evader crashes alone, and double-crash rate where both agents crash simultaneously.
Finally, the time-to-catch measures how long pursuers take to capture the evader.

Time-to-catch is biased downward because it only includes successful catches: weaker pursuers may seem faster by catching only easy targets without crashing.
We therefore use a right-censored metric, setting the time-to-catch to 10 seconds whenever an episode ends in a crash or timeout.

We train then evaluate the performances of our strategies in an arena of size $32\times32\times16$ meters, with the evader constrained in a smaller volume of size $12\times12\times6$ meters in the centre of the arena. For each combination of pursuers and evader strategies, we run 10,000 episodes and report the averaged metrics in Table~\ref{tab:performances}.

\subsection{Comparison with baselines}
\input{table.tex}

We compare the performances of the trained policies in the 3v1 scenario against both heuristic baselines and learning-based methods.
Many recent works have applied RL to multi-pursuer pursuit–evasion tasks \cite{Xiao2024, Zhang2023, Chen2024, zhang2023dacoop}, sometimes using advanced encoding mechanisms to predict the evader’s motion or account for obstacles \cite{Zhang2023, Chen2024}. 
However, they train against deterministic evaders. Only a few addresses high-speed manoeuvring drones, and most use velocity control. To our knowledge, none combines agile drone dynamics with a learning evader.
But, we believe that these learning methods are not incompatible with an SP framework, and could still perform well against learning evaders. 
However, our aim in this paper is to show that PFSP combined with thrust–body-rate control produces robust policies for agile pursuit–evasion.
Since evader state estimation in high-speed flight is outside our scope, we restrict learning-based comparisons to ablations that isolate the effect of PFSP and thrust-body-rate control on performances.
And we encourage future works to address the benchmarking of the existing RL-based pursuit-evasion methods in an SP framework.

In particular for the pursuers, we will compare our PFSP-CTBR policies with:
\begin{itemize}
  \item Fast-Response Proportional Navigation (FRPN) \cite{pliska2024towards}: an evolution of Proportional Navigation for manoeuvring multi-rotors. This method is designed for high-speed interception of single agile target with a single pursuer using acceleration commands. There is no inter-pursuer collision avoidance in the original implementation.
  \item Artificial Potential Field (APF): a method where the pursuers are attracted by the evader and repelled by the arena boundaries and other pursuers. We use the implementation from \cite{zhang2023dacoop} which uses the normalized sum of the attractive and repulsive forces as the direction of a constant velocity command.
  \item FRPN + APF: we combine the FRPN method with an APF-based collision avoidance, where the acceleration command from FRPN is summed with the repulsive forces from APF to obtain the final acceleration command.
  \item noSP: an ablation of our method where the pursuers are trained with MAPPO against a deterministic evader that moves away from the pursuers using APF navigation.
  \item SP: an ablation of our method where the pursuers are trained with MAPPO and SP, but face only the latest version of their opponent at each episode.
  \item PFSP-vel: an ablation of our method where the pursuers are trained with PFSP but output a desired velocity instead of thrust and body rates. The desired velocity is converted to thrust and body rates using an INDI controller described in Section~\ref{sec:methodology_training_setup}. 
\end{itemize} 
For the evader:
\begin{itemize}
  \item APF: the evader is repelled by the pursuers and the arena boundaries. We use the implementation from \cite{zhang2023dacoop}.
  \item noSP: an ablation of our method where the evader is trained with PPO against deterministic pursuers that use APF to intercept the evader.
  \item SP: an ablation of our method where the evader is trained with MAPPO and SP, but faces only the latest version of its opponent at each episode.
  \item PFSP-vel: an ablation of our method where the evader is trained with PFSP but outputs a desired velocity instead of thrust and body rates. The desired velocity is converted to thrust and body rates using an INDI controller described in Section~\ref{sec:methodology_training_setup}. 
\end{itemize}
For fairness, these methods only access the opponent’s position and velocity, as do our RL policies.


All the pursuers heuristic baselines performances drop when facing the agile learned evaders. 
FRPN crash rates is especially high, as it does not consider boundary constraints. This effect is exacerbated with multiple pursuers as they tend to collide into each other. 
APF repulsion from arena boundaries creates safer flights, but the simple attraction to the evader current position leads to a “Pure-Pursuit behaviour” with the pursuers chasing the target indefinitely never reaching it.
Combining FRPN with an APF repulsion leads to a higher catch rate, but the pursuers still fails to catch our learned evader consistently.   

In contrast, PFSP-CTBR, which was trained to avoid crashes, exhibits a much lower crash rate against the more agile learned evader. 
It achieves the highest catch rate and the lowest time-to-catch against all evaders, indicating that it learns effective interception strategies across diverse evasive manoeuvres while respecting arena boundaries and reducing inter-pursuer collisions. 
Its low time-to-catch further suggests that PFSP-CTBR succeeds even against the most challenging evaders, where other methods either fail or crash.
PFSP-CTBR also produces the strongest evader policy, with higher evade rates and lower crash rates than the other evaders across all settings.
This shows that it can consistently avoid capture for the full episode duration or until a pursuer crashes, the PFSP-CTBR pursuers strategy being the only one reliably able to catch it. 
It is particularly effective against PN pursuers, which tend to crash when facing agile maneuvers.

\subsubsection{Impact of control level}
The crash rate of the PFSP-vel evader increases when facing pursuers using CTBR or acceleration control. 
Although PFSP-vel pursuers maintain strong catch rates against most evaders, they are ineffective against the PFSP-CTBR evader. 
Across all settings, PFSP-CTBR achieves higher catch and evade rates while also reducing crash rates compared with PFSP-vel. 
These results indicate that velocity control constrains agility, whereas CTBR supports safer and more robust policies by enabling more agile maneuvers. 
This advantage is particularly pronounced for the PFSP-CTBR evader, which consistently combines high evade rates with low crash rates.

\subsubsection{Effect of Self-Play}
The no-SP strategies show strong performances against the specific opponents encountered during training: 
the pursuers reliably capture the APF evader, while the evader achieves a high evade rate against FRPN-APF, which as a result crashes a lot.
But against other learned evaders, the no-SP pursuers degrades substantially, with crash rates rising sharply against the most agile policies. 
This indicates overfitting to deterministic training opponents and poor generalization to unseen strategies.

\begin{figure*}[!ht]
  \centering
  \includegraphics[width=\textwidth, clip, trim=0 15 0 15]{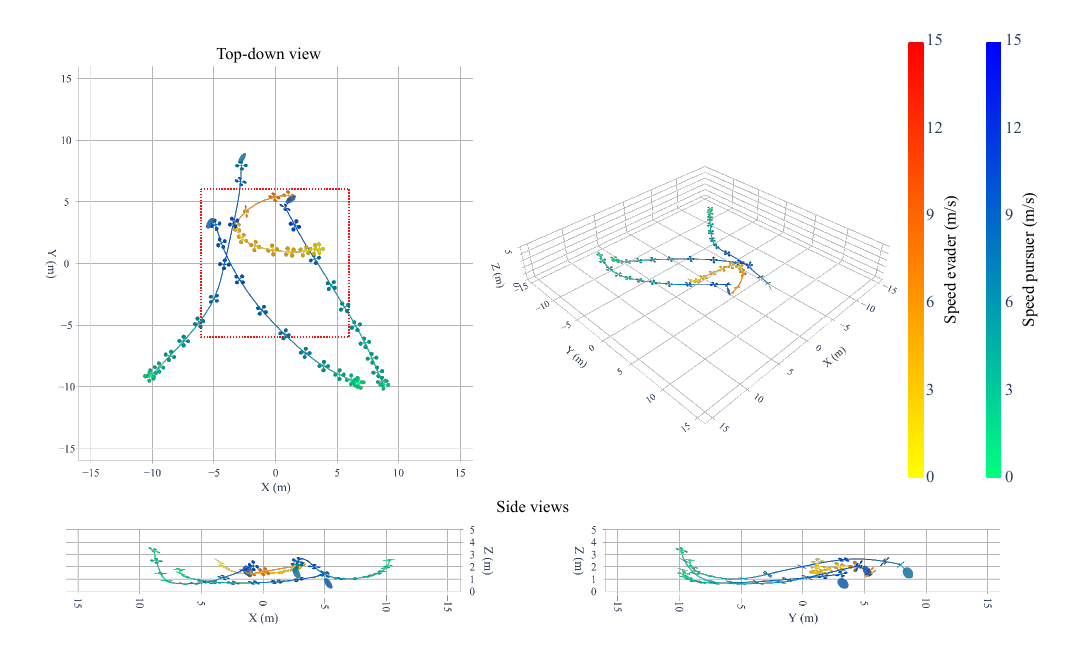}
  \caption{A successful capture: The three pursuers split up and push the evader towards the inner boundaries, blocking its escape routes and anticipating that it will avoid crossing them in order to capture it.}
  \label{fig:3v1_capture}
\end{figure*}

This issue is less pronounced for SP policies.
Their catch rate stays high against all evaders except PFSP-CTBR, even against PFSP-vel.
And their evade and crash rates remain consistent against all pursuers other than PFSP-CTBR.
Training against agile opponents appears to produce strategies that are more robust to unpredictable trajectories and still effective against less agile adversaries. The 3v1 setting favors the pursuers by reducing the evader’s escape routes, and likely alleviates the issue of overfitting against the last opponent seen during training.
In comparison, the PFSP-CTBR agents have uniformly strong performance.  
By prioritizing challenging opponents during training, it maintains a stronger learning signal and produces stronger strategies, that also generalize better to unseen opponents.


\subsection{Qualitative Results}

\begin{figure}[t]
  \begin{subfigure}[t]{\linewidth}
    \centering
    \includegraphics[width=\linewidth, clip, trim=0 100 0 195]{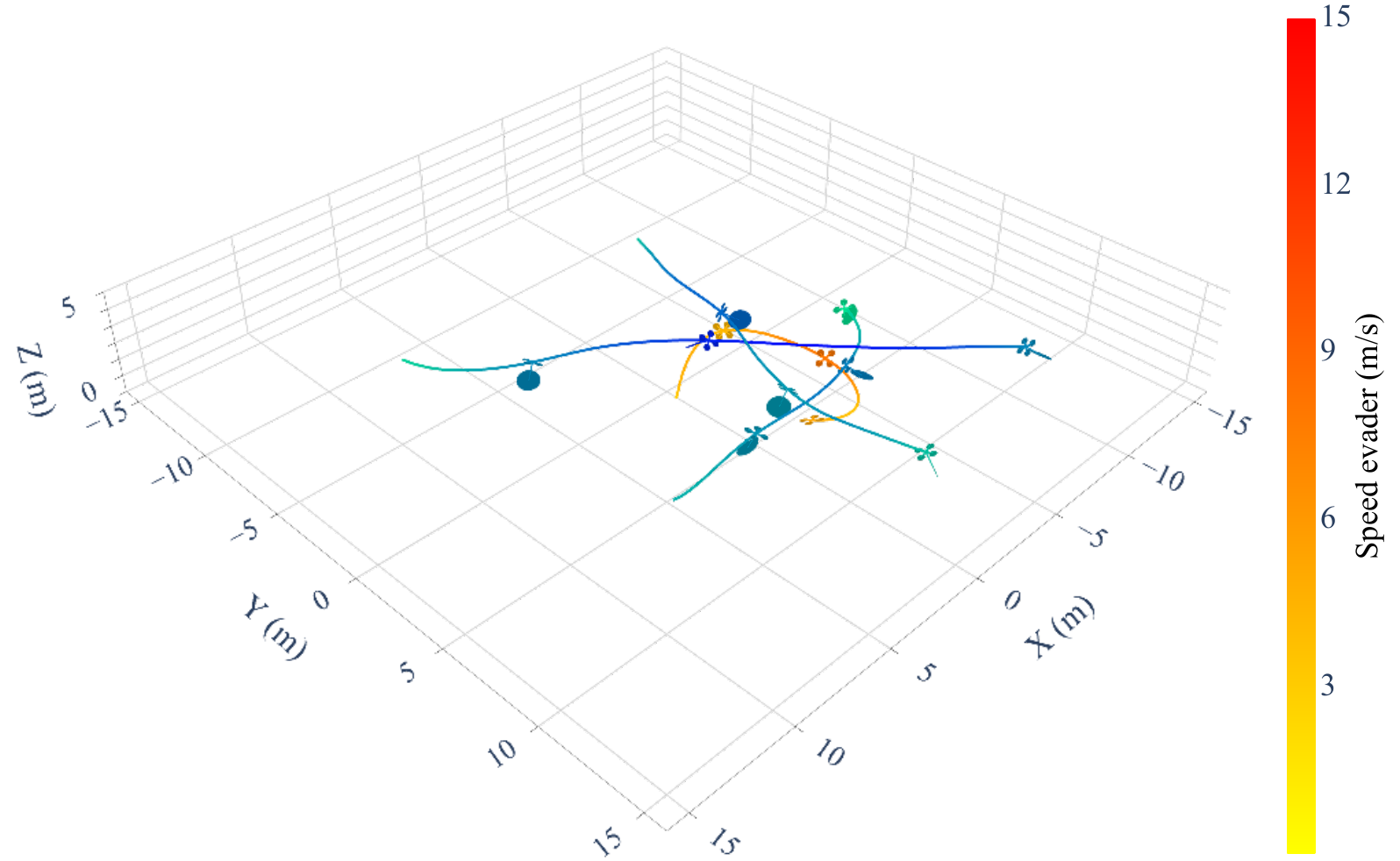}
    \caption{}
    \label{fig:3v1_slipby}
  \end{subfigure}
  \begin{subfigure}[b]{\linewidth}
    \centering
    \includegraphics[width=\linewidth, clip, trim=0 100 0 195]{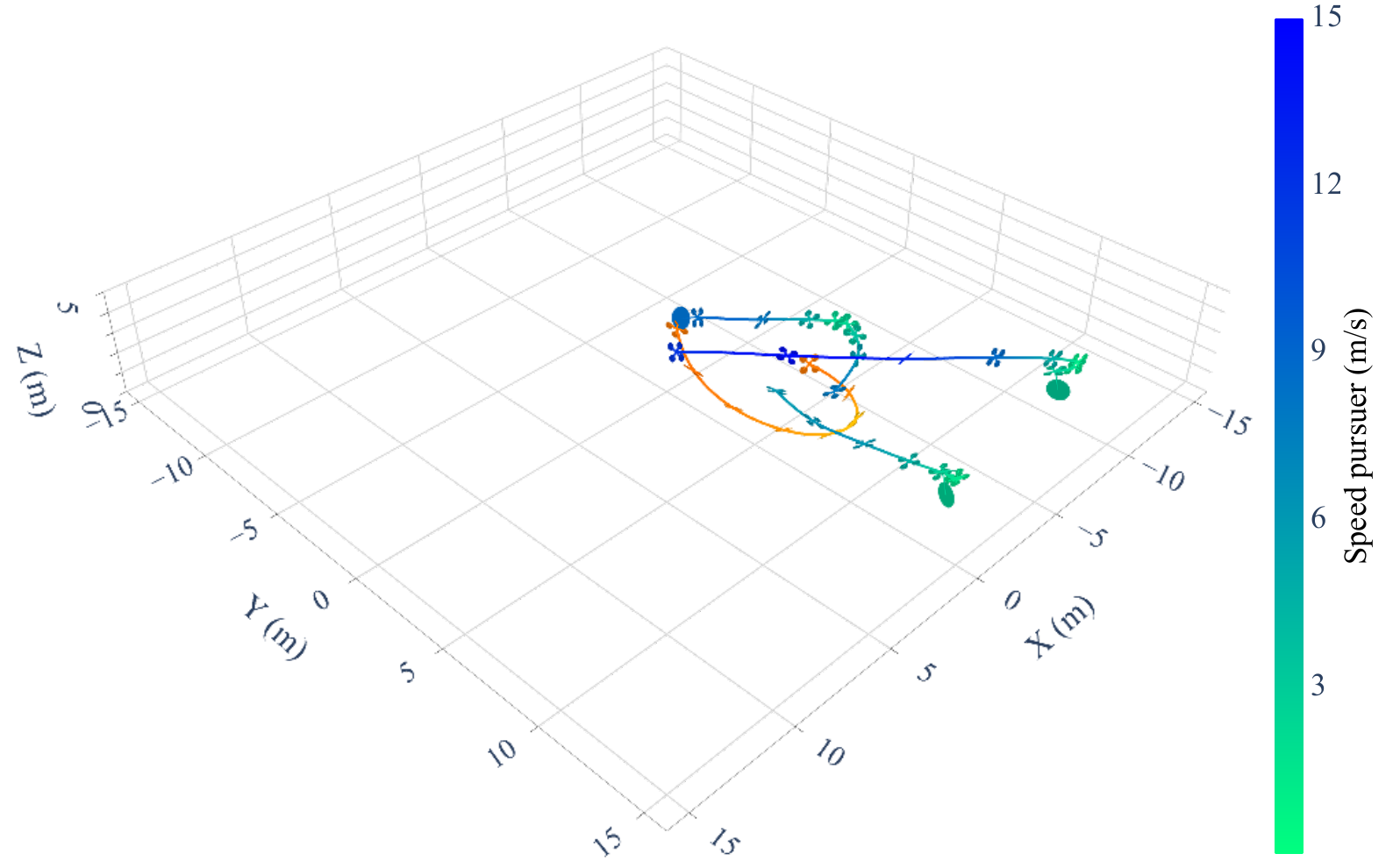}
    \caption{}
    \label{fig:3v1_retry}
  \end{subfigure}
  \caption{Two successive interception attempts: (a) the evader narrowly avoids capture by the three pursuers through a swift 3D manoeuvre, (b) after the evader slips by, the pursuers quickly adapt their strategy and one pursuer makes a rapid second attempt to successfully intercept the evader.}
  \label{fig:miss_and_recover}
\end{figure}

In this section we present qualitative results of our trained pursuers and evader in simulation.
Similar to prior work on the 1v1 case \cite{gavinagile}, the evader learned a diverse set of agile evasive manoeuvres that use the full 3D space to avoid capture, such evasive manoeuvres can be seen in Fig. \ref{fig:3v1_slipby}.
In response, the pursuers learned to anticipate these manoeuvres and to maximize the surface of the catching net facing the evader to increase the chances of a successful capture. 

\begin{figure}[t]
  \begin{subfigure}[t]{0.49\linewidth}
    \centering
    \includegraphics[width=\linewidth, clip, trim=50 20 50 10]{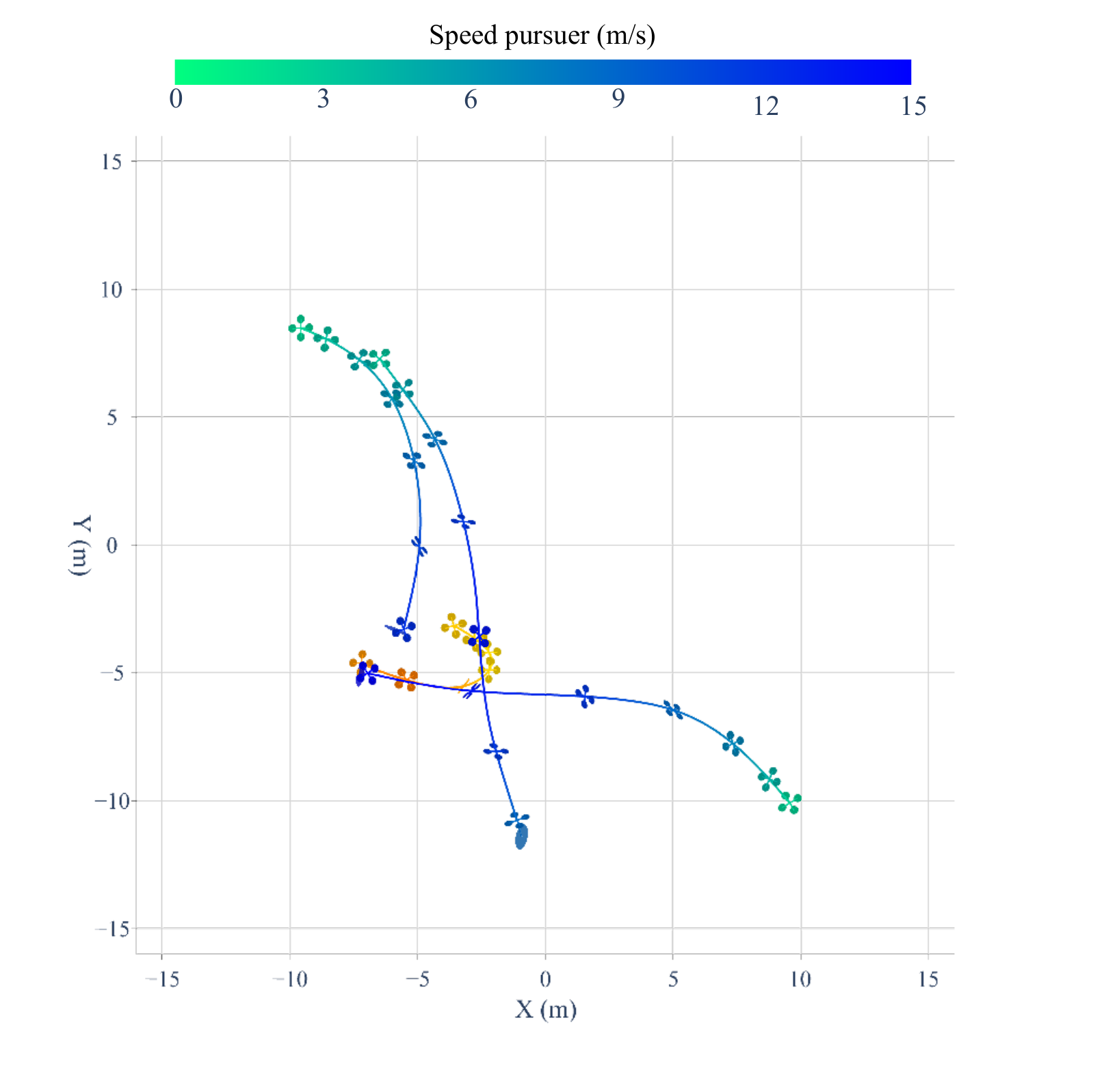}
    \caption{}
    \label{fig:speed_up}
  \end{subfigure}
  \hfill
  \begin{subfigure}[t]{0.49\linewidth}
    \centering
    \includegraphics[width=\linewidth, clip, trim=50 20 50 10]{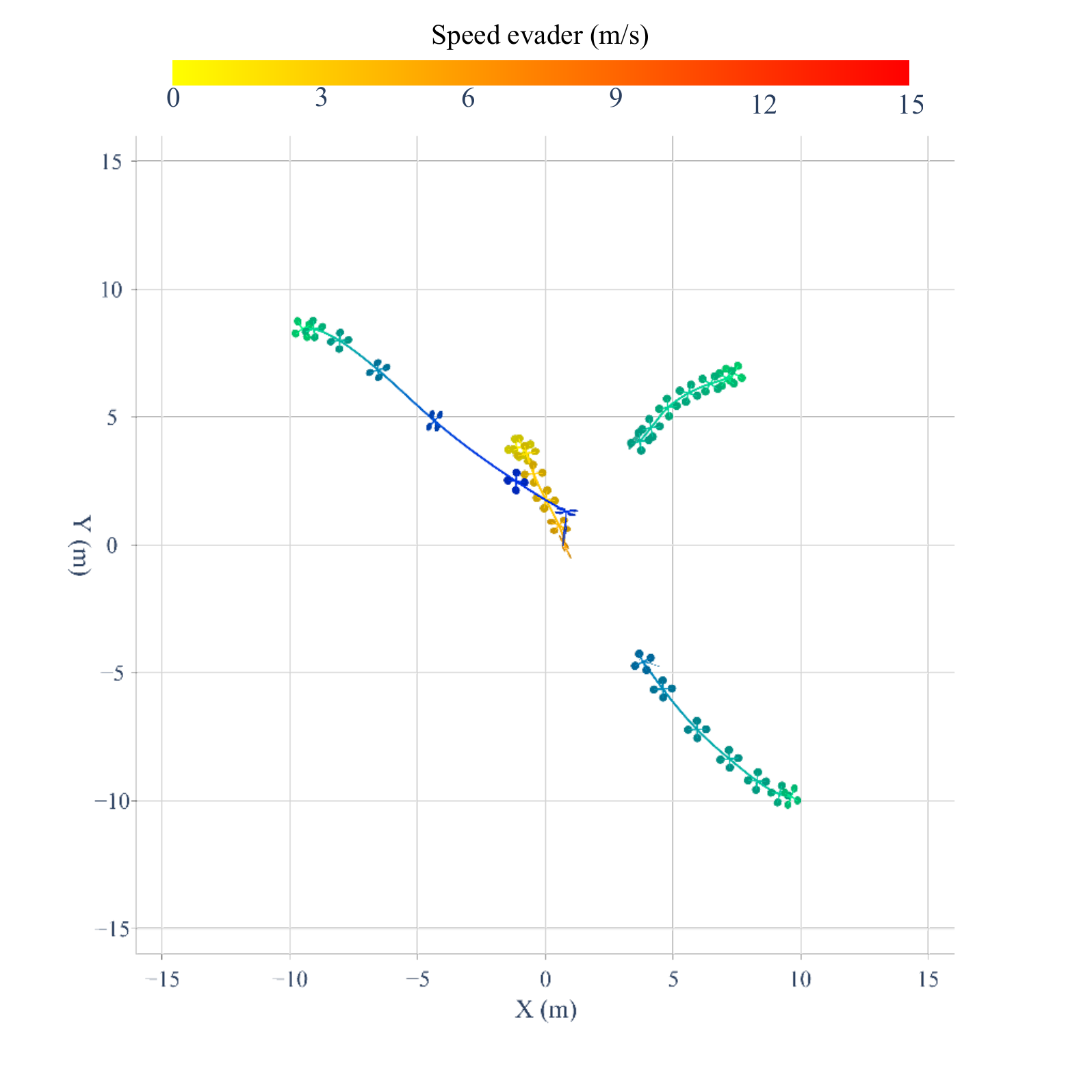}
    \caption{}
    \label{fig:3v1_encircle}
  \end{subfigure}
  \caption{Two additional examples of the pursuers' coordinated strategies: in (a) the pursuers coordinate their interception attempts, one pursuer speeds up to make a first attempt while the two others approach slower to wait for the opportunity of a second attempt if the first one fails. In (b) the pursuers exploit an encircling position to approach from all directions and capture the target.}
\end{figure}

They also learned to exploit their number to increase catching chances.
The pursuers learned to avoid inter-agent crashes and split to attempts at capture in close succession to limit the risk of collisions. One pursuer speed up to make a first capture attempt, while the two others approach slower to wait for the opportunity of a second attempt if the first one fails, which is what happens in Fig. \ref{fig:speed_up}.

The number of episodes where the pursuers adopt an encircling strategy to surround the evader and reduce its escape routes is low.
The pursuers seem to have learned to rely more on their agility to attempt coordinated successive rapid interception attempts rather than to rely on a more strategic encircling strategy.
During early training stages, the evader strategy being simpler and less agile could have led the pursuers to first adopt more reactive and direct interception strategies, that are slow to be replaced by more strategic encircling later in the training stages. 
This could also be explained by the evader being constrained in a smaller volume in the centre of the arena, which make it easier to catch without the need for encircling strategies as the evader has already limited escape routes. A bigger arena could alleviate this issue.
On the contrary, the pursuers learned well to exploit the evader's reluctance to approach its constrained boundaries. They will split up and push it towards the edges, blocking its escape routes. They will then anticipate the evader avoiding crossing them to capture it. This behaviour is shown in Fig. \ref{fig:3v1_capture}.  
Still, when in favourable positions, the pursuers did learn to profit of the encircling positions to approach from all directions and capture the target as shown in Fig. \ref{fig:3v1_encircle}.

Interestingly, the pursuers exhibit adaptive behaviour, quickly recovering from failed interception attempts and making multiple successive tries within the allowed capture time. As illustrated in Fig.~\ref{fig:3v1_retry}, after the evader narrowly escapes in Fig.~\ref{fig:3v1_slipby}, one pursuer rapidly re-engages and successfully completes the interception.

In the rare case when the pursuers attempt a capture alone, it usually results in a failure as the evader can escape easily.
\section{Conclusions}\label{sec:conclusions}
 
In this paper, we presented a Deep RL approach for the pursuit-evasion task with multiple pursuers and one evader, using MARL and PFSP to train high-speed manoeuvring policies for both teams.
Our method achieves a high capture rate and low time to catch against a variety of agile evader strategies, while maintaining low crash rates for both pursuers and evaders. 
We show that our method outperforms several heuristic baselines in simulation achieving higher catch rates and lower crash rates when facing agile evaders. 
Ablation studies demonstrated the importance of self-play and the control level for learning robust policies in this task.
Qualitative analysis of our trained policies reveals emergent behaviours, such as coordinated interception attempts and adaptive recovery from failed attempts.
Furthermore, our simulation environment and training loop being fully within the JAX framework was essential, as it enabled massively parallelized execution and drastically reduced training times, making extensive RL training computationally feasible.

While our approach shows promising results,  
policies are evaluated only in simulation and may behave differently in real-world settings due to sensor noise, unmodeled dynamics, and disturbances. Future work will transfer them to real quadrotors and incorporate perception and state estimation for practical deployment.

\addtolength{\textheight}{0cm}   




\bibliographystyle{IEEEtran} 
\bibliography{biblio}


\end{document}

%% file: table.tex
\begin{table}[!htbp] 
  \centering
  \caption{Performances of the pursuers and the evader.}
  \label{tab:performances}
  \begin{tabular}{@{}rr@{\hskip 9pt}rrrrr} \toprule
    \multicolumn{2}{c}{\textbf{Pursuer mode}}
                                                               & \multicolumn{5}{c}{\textbf{Evader Mode}}                             \\
    \midrule
    \multicolumn{2}{c}{\textbf{FRPN \cite{pliska2024towards}}} & \textbf{APF}          & \textbf{noSP}         & \textbf{SP}            & \textbf{PFSP-vel}       & \textbf{PFSP}                 \\                 
    \cmidrule[0.01em](l){1-7} \noalign{\vskip -\belowrulesep}                                         
    \multicolumn{2}{r@{\hskip 9pt}}{\rule{0pt}{2.2ex}\rule{0pt}{2.2ex}\textbf{Catch Rate (\%)}}               & \gH{55.0}  & \gH{46.8}  & \gH{18.2}   & \gH{22.2}    & \gH{2.9}                       \\                 
    \multicolumn{2}{r@{\hskip 9pt}}{\textbf{Evade Rate (\%)}}               & \gL{0.1}   & \gL{4.5}   & \gL{16.2}   & \gL{6.8}     & \gL{27.1}                    \\                 
    \multicolumn{2}{r@{\hskip 9pt}}{\textbf{Crash Rate (\%)}}               & \gL{44.9}  & \gL{48.7}  & \gL{65.6}   & \gL{71.0}    & \gL{69.9}                    \\[0.0em]                          
                                  & Pursuer                    & \gLs{37.1}     & \gLs{47.8}     & \gLs{64.3}      & \gLs{54.2}       & \gLs{69.7}             \\[-0.15em]                        
                                  & Evader                     & \gLs{5.5}      & \gLs{0.0}      & \gLs{0.0}       & \gLs{16.4}       & \gLs{0.2}              \\[-0.15em]                           
                                  & Double                     & \gLs{2.3}      & \gLs{0.9}      & \gLs{1.3}       & \gLs{0.4}        & \gLs{0.1}\\                 
                    \multicolumn{2}{r@{\hskip 9pt}}{\rule{0pt}{2.2ex}\textbf{Time to Catch (s)}}             & \gT{5.75}  & \gT{6.64}  & \gT{8.75}  & \gT{8.51}  & \gT{9.99}    \\[0.0em]                 
                                                              &±std& \gTs{5.75}{(±4.18)}  & \gTs{6.64}{(±4.04)}  & \gTs{8.75}{(±3.22)}   & \gTs{8.51}{(±3.37)}    & \gTs{9.99}{(±1.16)}          \\                 
    \noalign{\vskip -\aboverulesep}
    \midrule                                                   
    \multicolumn{2}{c}{\textbf{APF}}                           & \textbf{APF}          & \textbf{noSP}         & \textbf{SP}            & \textbf{PFSP-vel}       & \textbf{PFSP}                 \\                 
    \cmidrule[0.01em](l){1-7} \noalign{\vskip -\belowrulesep}                                          
    \multicolumn{2}{r@{\hskip 9pt}}{\rule{0pt}{2.2ex}\textbf{Catch Rate (\%)}}               & \gH{50.9}  & \gH{2.4}   & \gH{8.9}    & \gH{1.0}     & \gH{0.5}                       \\                 
    \multicolumn{2}{r@{\hskip 9pt}}{\textbf{Evade Rate (\%)}}               & \gL{17.7}  & \gL{93.9}  & \gL{88.2}   & \gL{86.9}    & \gL{96.6}                    \\                 
    \multicolumn{2}{r@{\hskip 9pt}}{\textbf{Crash rates (\%)}}              & \gL{31.4}  & \gL{3.8}   & \gL{2.9}    & \gL{12.0}    & \gL{3.0}                       \\[0.0em]                        
                                  & Pursuer                    & \gLs{14.5}     & \gLs{3.7}      & \gLs{2.6}       & \gLs{3.0}        & \gLs{1.8}              \\[-0.15em]                        
                                  & Evader                     & \gLs{15.4}     & \gLs{0.0}      & \gLs{0.0}       & \gLs{9.0}        & \gLs{1.2}              \\[-0.15em]                        
                                  & Double                     & \gLs{1.5}      & \gLs{0.0}      & \gLs{0.2}       & \gLs{0.0}        & \gLs{0.0}\\                 
                    \multicolumn{2}{r@{\hskip 9pt}}{\rule{0pt}{2.2ex}\textbf{Time to Catch (s)}}             & \gT{7.36}  & \gT{9.99}  & \gT{9.58}  & \gT{9.99}  & \gT{10.00}   \\
                                                                &±std& \gTs{7.36}{(±3.32)}  & \gTs{9.99}{(±1.26)}  & \gTs{9.58}{(±2.19)}   & \gTs{9.99}{(±0.93)}    & \gTs{10.00}{(±0.52)}           \\                 
    \noalign{\vskip -\aboverulesep}
    \midrule                                                   
    \multicolumn{2}{c}{\textbf{FRPN + APF}}                    & \textbf{APF}          & \textbf{noSP}         & \textbf{SP}            & \textbf{PFSP-vel}       & \textbf{PFSP}               \\                 
    \cmidrule[0.01em](l){1-7} \noalign{\vskip -\belowrulesep}                                          
    \multicolumn{2}{r@{\hskip 9pt}}{\rule{0pt}{2.2ex}\textbf{Catch Rate (\%)}}               & \gH{71.9}  & \gH{47.0}  & \gH{64.7}   & \gH{59.6}    & \gH{20.0}                    \\                 
    \multicolumn{2}{r@{\hskip 9pt}}{\textbf{Evade Rate (\%)}}               & \gL{0.2}   & \gL{11.7}  & \gL{16.4}   & \gL{9.8}     & \gL{40.2}                    \\                 
    \multicolumn{2}{r@{\hskip 9pt}}{\textbf{Crash rates (\%)}}              & \gL{27.9}  & \gL{41.3}  & \gL{18.9}   & \gL{30.6}    & \gL{39.8}                    \\[0.0em]                        
                                  & Pursuer                    & \gLs{21.4}     & \gLs{40.6}     & \gLs{18.2}      & \gLs{13.8}       & \gLs{39.5}             \\[-0.15em]                        
                                  & Evader                     & \gLs{4.6}      & \gLs{0.0}      & \gLs{0.0}       & \gLs{16.6}       & \gLs{0.3}              \\[-0.15em]                        
                                  & Double                     & \gLs{1.8}      & \gLs{0.7}      & \gLs{0.7}       & \gLs{0.2}        & \gLs{0.0}\\    
                          \multicolumn{2}{r@{\hskip 9pt}}{\rule{0pt}{2.2ex}\textbf{Time to Catch (s)}}             & \gT{4.83}  & \gT{7.0}   & \gT{5.87}  & \gT{6.22}  & \gT{9.14}    \\
                                                                  &±std& \gTs{4.83}{(±3.55)} & \gTs{7.0}{(±3.75)}  & \gTs{5.87}{(±3.77)}   & \gTs{6.22}{(±3.76)}    & \gTs{9.14}{(±2.47)}          \\                 
    \noalign{\vskip -\aboverulesep}
    \midrule                                                   
    \multicolumn{2}{c}{\textbf{no-SP}}                         & \textbf{APF}          & \textbf{noSP}         & \textbf{SP}            & \textbf{PFSP-vel}       & \textbf{PFSP}                 \\                 
    \cmidrule[0.01em](l){1-7} \noalign{\vskip -\belowrulesep}                                          
    \multicolumn{2}{r@{\hskip 9pt}}{\rule{0pt}{2.2ex}\textbf{Catch Rate (\%)}}               & \gH{65.2}  & \gH{43.9}  & \gH{23.0}   & \gH{20.2}    & \gH{3.5}                     \\                 
    \multicolumn{2}{r@{\hskip 9pt}}{\textbf{Evade Rate (\%)}}               & \gL{5.7}   & \gL{24.2}  & \gL{11.9}   & \gL{9.8}     & \gL{11.9}                    \\                 
    \multicolumn{2}{r@{\hskip 9pt}}{\textbf{Crash rates (\%)}}              & \gL{29.1}  & \gL{31.9}  & \gL{65.1}   & \gL{70.0}    & \gL{84.6}                    \\[0.0em]                        
                                  & Pursuer                    & \gLs{11.9}     & \gLs{31.2}     & \gLs{64.4}      & \gLs{42.6}       & \gLs{84.3}             \\[-0.15em]                        
                                  & Evader                     & \gLs{15.1}     & \gLs{0.2}      & \gLs{0.0}       & \gLs{27.0}       & \gLs{0.2}              \\[-0.15em]                        
                                  & Double                     & \gLs{2.1}      & \gLs{0.5}      & \gLs{0.7}       & \gLs{0.4}        & \gLs{0.1}\\                 
                    \multicolumn{2}{r@{\hskip 9pt}}{\rule{0pt}{2.2ex}\textbf{Time to Catch (s)}}             & \gT{5.49}  & \gT{7.39}  & \gT{8.68}  & \gT{8.80}  & \gT{9.99}    \\
                                                                  &±std& \gTs{5.49}{(±3.96)}  & \gTs{7.39}{(±3.64)}  & \gTs{8.68}{(±3.05)}   & \gTs{8.80}{(±3.02)}    & \gTs{9.99}{(±1.38)}          \\                 
    \noalign{\vskip -\aboverulesep}
    \midrule                                                   
    \multicolumn{2}{c}{\textbf{SP}}                            & \textbf{APF}          & \textbf{noSP}         & \textbf{SP}            & \textbf{PFSP-vel}       & \textbf{PFSP}                 \\                 
    \cmidrule[0.01em](l){1-7} \noalign{\vskip -\belowrulesep}                                          
    \multicolumn{2}{r@{\hskip 9pt}}{\rule{0pt}{2.2ex}\textbf{Catch Rate (\%)}}               & \gH{69.3}  & \gH{62.5}  & \gH{50.9}   & \gH{49.4}    & \gH{22.5}                    \\                 
    \multicolumn{2}{r@{\hskip 9pt}}{\textbf{Evade Rate (\%)}}               & \gL{9.0}   & \gL{22.1}  & \gL{25.4}   & \gL{16.3}    & \gL{45.0}                    \\                 
    \multicolumn{2}{r@{\hskip 9pt}}{\textbf{Crash rates (\%)}}              & \gL{21.7}  & \gL{15.4}  & \gL{23.7}   & \gL{34.3}    & \gL{32.5}                    \\[0.0em]                        
                                  & Pursuer                    & \gLs{9.0}      & \gLs{14.9}     & \gLs{22.5}      & \gLs{16.4}       & \gLs{32.1}             \\[-0.15em]                        
                                  & Evader                     & \gLs{10.6}     & \gLs{0.3}      & \gLs{0.0}       & \gLs{17.6}       & \gLs{0.2}              \\[-0.15em]                        
                                  & Double                     & \gLs{2.0}      & \gLs{0.2}      & \gLs{1.2}       & \gLs{0.3}        & \gLs{0.2}\\                 
    \multicolumn{2}{r@{\hskip 9pt}}{\textbf{Time to Catch (s)}}             & \gT{5.07}  & \gT{5.63}  & \gT{6.73}  & \gT{6.60}  & \gT{8.70}    \\                 
                                                                  &±std& \gTs{5.07}{(±3.93)}  & \gTs{5.63}{(±4.03)}  & \gTs{6.73}{(±3.86)}   & \gTs{6.60}{(±4.00)}    & \gTs{8.70}{(±3.10)}          \\                 
    \noalign{\vskip -\aboverulesep}
    \midrule                                                   
    \multicolumn{2}{c}{\textbf{PFSP-vel}}                      & \textbf{APF}          & \textbf{noSP}         & \textbf{SP}            & \textbf{PFSP-vel}       & \textbf{PFSP}               \\                 
    \cmidrule[0.01em](l){1-7} \noalign{\vskip -\belowrulesep}                                          
    \multicolumn{2}{r@{\hskip 9pt}}{\rule{0pt}{2.2ex}\textbf{Catch Rate (\%)}}               & \gH{59.0}  & \gH{70.8}  & \gH{63.8}   & \gH{67.2}    & \gH{17.7}                    \\                 
    \multicolumn{2}{r@{\hskip 9pt}}{\textbf{Evade Rate (\%)}}               & \gL{17.2}  & \gL{24.1}  & \gL{32.3}   & \gL{20.0}    & \gL{77.6}                    \\                 
    \multicolumn{2}{r@{\hskip 9pt}}{\textbf{Crash rates (\%)}}              & \gL{23.8}  & \gL{5.1}   & \gL{3.9}    & \gL{12.9}    & \gL{4.7}                     \\[0.0em]                        
                                  & Pursuer                    & \gLs{10.8}     & \gLs{4.7}      & \gLs{3.6}       & \gLs{8.3}        & \gLs{4.5}              \\[-0.15em]                        
                                  & Evader                     & \gLs{12.7}     & \gLs{0.0}      & \gLs{0.0}       & \gLs{4.4}        & \gLs{0.1}              \\[-0.15em]                        
                                  & Double                     & \gLs{0.3}      & \gLs{0.3}      & \gLs{0.3}       & \gLs{0.2}        & \gLs{0.1}\\  
                    \multicolumn{2}{r@{\hskip 9pt}}{\rule{0pt}{2.2ex}\textbf{Time to Catch (s)}}             & \gT{6.33}  & \gT{5.68}  & \gT{6.56}  & \gT{6.01}  & \gT{9.30}    \\
                                                                  &±std& \gTs{6.33}{(±3.72)}  & \gTs{5.68}{(±3.61)}  & \gTs{6.56}{(±3.45)}   & \gTs{6.01}{(±3.59)}    & \gTs{9.30}{(±2.33)}          \\                 
    \noalign{\vskip -\aboverulesep}
    \midrule                                                   
    \multicolumn{2}{c}{\textbf{PFSP-CTBR}}                     & \textbf{APF}          & \textbf{noSP}         & \textbf{SP}            & \textbf{PFSP-vel}       & \textbf{PFSP}               \\                 
    \cmidrule[0.01em](l){1-7} \noalign{\vskip -\belowrulesep}                                          
    \multicolumn{2}{r@{\hskip 9pt}}{\rule{0pt}{2.2ex}\textbf{Catch Rate (\%)}}               & \gH{88.1}  & \gH{91.3}  & \gH{93.3}   & \gH{76.9}    & \gH{88.4}                    \\                 
    \multicolumn{2}{r@{\hskip 9pt}}{\textbf{Evade Rate (\%)}}               & \gL{0.2}   & \gL{0.3}   & \gL{0.3}    & \gL{0.2}     & \gL{3.4}                     \\                 
    \multicolumn{2}{r@{\hskip 9pt}}{\textbf{Crash rates (\%)}}              & \gL{11.8}  & \gL{8.4}   & \gL{6.4}    & \gL{22.9}    & \gL{8.2}                     \\[0.0em]                        
                                  & Pursuer                    & \gLs{4.1}      & \gLs{7.9}      & \gLs{4.0}       & \gLs{8.2}        & \gLs{7.7}              \\[-0.15em]                        
                                  & Evader                     & \gLs{4.9}      & \gLs{0.0}      & \gLs{0.0}       & \gLs{14.3}       & \gLs{0.3}              \\[-0.15em]                        
                                  & Double                     & \gLs{2.8}      & \gLs{0.4}      & \gLs{2.4}       & \gLs{0.4}        & \gLs{0.2}\\                 
                    \multicolumn{2}{r@{\hskip 9pt}}{\rule{0pt}{2.2ex}\textbf{Time to Catch (s)}}             & \gT{3.06}  & \gT{2.85}  & \gT{2.82}  & \gT{4.10}  & \gT{3.71}    \\
      &±std& \gTs{3.06}{(±2.95)}  & \gTs{2.85}{(±2.70)}  & \gTs{2.82}{(±2.41)}   & \gTs{4.10}{(±3.61)}    & \gTs{3.71}{(±2.96)}          \\                 
    [-0.2em]\bottomrule
    \multicolumn{7}{l}{%
    \scriptsize \textbf{Legend:} Cell colours indicate performance from the pursuer's perspective}\\[-0.1em]
    \multicolumn{7}{l}{%
    \begin{tabular}{@{}lc@{}}
      \scriptsize Catch rate (\%) &
      \GradientLegendStrip[5.0cm]
      {0.0}{50.0}{100.0}
      {heatbad}{heatmid}{heatgood}{\opacity}
      {}
      \\[-0.3em]
      \scriptsize Evade/crash rates (\%) &
      \GradientLegendStrip[5.0cm]
      {0.0}{30.0}{100.0}
      {heatgood}{heatmid}{heatbad}{\opacity}
      {}
      \\[-0.3em]
      \scriptsize Time to catch (s) &
      \GradientLegendStrip[5.0cm]
        {2.82}{6.41}{10.00}
        {heatgood}{heatmid}{heatbad}{\opacity}
        {}
    \end{tabular}%
    }\\[-0.3em]

  \end{tabular}
\end{table}